\pdfoutput=1
\documentclass[11pt]{article}

\usepackage[preprint]{acl}

\usepackage{times}
\usepackage{latexsym}

\usepackage[T1]{fontenc}
\usepackage[utf8]{inputenc}

\usepackage{microtype}

\usepackage{inconsolata}

\usepackage{graphicx}

\usepackage{multirow}
\usepackage{tabularx}
\usepackage{xurl}
\usepackage{hyperref}
\usepackage{array} % for custom column types
\usepackage[ruled,vlined,linesnumbered,noend]{algorithm2e}
\SetKwComment{Comment}{$\triangleright$\ }{}
\SetAlgoLined
\usepackage{amsmath}
\newcolumntype{P}[1]{>{\raggedright\arraybackslash}p{#1}} % ragged-right paragraph col
\usepackage{listings}
\usepackage{listingsutf8} % <--- important for UTF-8 in listings

\title{Do You Get the Hint? Benchmarking LLMs on the Board Game Concept}

\author{
  \textbf{Ine Gevers} \and \textbf{Walter Daelemans}
\\
  CLiPS, University of Antwerp
\\
  \small{
    \href{mailto:ine.gevers@uantwerpen.be}{ine.gevers@uantwerpen.be},\;
    \href{mailto:walter.daelemans@uantwerpen.be}{walter.daelemans@uantwerpen.be}
  }
}

\begin{document}
\maketitle
\begin{abstract}
Large language models (LLMs) have achieved striking successes on many benchmarks, yet recent studies continue to expose fundamental weaknesses.
In this paper, we introduce Concept, a simple word-guessing board game, as a benchmark for probing abductive reasoning. 
Our results show that this game, easily solved by humans (with a success rate of over 90\%), is still very challenging for state-of-the-art LLMs (no model exceeds 40\% success rate). Specifically, we observe that LLMs struggle with interpreting other players' strategic intents, and with correcting initial hypotheses given sequential information updates.
In addition, we extend the evaluation across multiple languages, and find that the LLM performance drops further in lower-resource languages (Dutch, French, and Spanish) compared to English. 
\end{abstract}

\section{Introduction}
\begin{figure}
    \centering
    \includegraphics[width=1\linewidth]{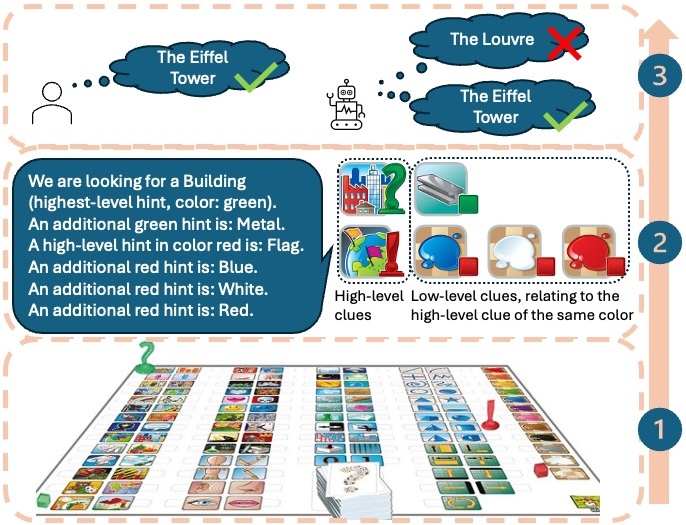}
    \caption{The core components of Concept. Step 1: we collect human game logs. Step 2: we extract the clues from Player 1 to create the prompts, which are then given to the LLM. In Step 3 we compare the performance between humans and LLMs on the task.} 
    \label{fig:example-concept}
\end{figure}

While Large Language models have propagated impressive progress in NLP recently, it remains a point of debate how much knowledge they possess about the world, and how (or if at all) they reason about this knowledge (e.g. \citet{chollet2025arc,kaesberg2025sparc,ngo2025mscore,benchekroun2023worldsense,dentella2024testing,gevers-etal-2025-benchmarks}). A creative way to assess such capabilities is to have LLMs play (natural language) games. In this paper, we introduce the boardgame Concept as a task to measure LLMs' abstract and abductive reasoning skills. We illustrate the construction of our benchmark in Figure \ref{fig:example-concept}. 
Similar to a single-modal game of Pictionary, there are two roles in Concept: \\

\textbf{Player 1}, the `clue-giver', chooses a concept (a word or a short phrase) and selects clues from a predefined list to express this concept to Player 2. Clues are added by using a pawn or cube of a specific color: clues of the same color are (hierarchically) related to each other. The first clue establishes the main category of the concept and is marked by a green question mark pawn (see Figure \ref{fig:example-concept}). Player 1 can give clusters of clues that relate to this main category by introducing exclamation mark pawns of different colors. The low-level clues (cubes) help to clarify the high-level clue (pawn) of the same color. Thus, low-level clues should be interpreted in relation to the high-level clue of that same color. For instance, in the example in Figure \ref{fig:example-concept}, the green question mark indicates the highest-level category (`Building'), where the small green hint ('Metal') refers to the material of the building. The red exclamation mark marks a high-level clue describing that building (`Flag'), and the low-level red clues (`Blue', `White', and `Red') should be understood as properties of that flag, not of the building directly. In this case, the clues mean `a metal building located in France'. 

\textbf{Player 2}, the `guesser', attempts to guess the concept based on the clues that are given. In a game, there can be multiple guessers. The only feedback Player 2 gets, is whether their answer is correct or not. Player 1 can in turn update their clues based on the incorrect answers, until Player 2 finds the correct answer.\\

We introduce Concept as a benchmark to evaluate abductive reasoning in LLMs. In this game, Player 1 selects clues to maximize the probability of the receiver correctly identifying the concept, while constrained to the set of available clues (118 unique items)\footnote{The icons with their textual representation can be found in the rule book, e.g., \url{https://shorturl.at/pWW0D}}. Player 2, in turn, has to use the available (incomplete) and abstract evidence (i.e., the clues) to efficiently search over a very open semantic space, to infer the most plausible concept that best satisfies all of the clues. To do so, Player 2 must rely on their own implicit background knowledge, which helps them to correctly interpret the clues (for instance, in the example in Figure \ref{fig:example-concept}, knowing that the clue `Metal' in relation to `Building' refers to the material the building is made of).
Since multiple answers could theoretically be correct given the clues, the player must also take into account the strategic intent of the clue-giver (relating to their Theory of Mind), and learn from previous incorrect answers.

Because the clue-giver has the freedom to express the concept however they want (while constrained to the available clues), this process of clue-giving is inherently more imprecise. This complicates a fair evaluation of LLMs (i.e., does the model not know the concept, or is it unable to select and combine seemingly unrelated clues in a meaningful way?). Therefore, in our setup, we implement LLMs solely as the guesser, using human-generated clues as input.\\
The effectiveness of the clues relies on how well the sender’s intended meaning aligns with the receiver’s interpretation. The players must converge on a shared understanding of the concept based on the limited, symbolic clues. This highlights the importance of not just linguistic comprehension, but also the strategic intent behind clue selection and interpretation. In order to perform well in Concept, it is not enough for the model to understand each clue separately, but it must be able to meaningfully combine these clues, within the shared conceptual framework that humans use. \\
Evaluating an LLM's performance on Concept thus assesses its capacity to participate such strategic linguistic interaction, relying on abductive reasoning that integrates associative skills (i.e., to disambiguate and connect the clues), Theory of Mind (i.e., to infer the strategic intention of Player 1), and common sense (i.e., to supply the (often unstated) background facts and everyday associations needed to correctly interpret the clues).

Concept has the advantage of being a very straightforward task, for which no expert knowledge is needed. Additionally, our setup allows us to compare performance with a human baseline, and the inclusion of multiple languages offers insights into the transferability of an LLM's abduction capabilities.

We collect game logs of finished games in English, French, Spanish, and Dutch from the online board game platform Board Game Arena\footnote{\url{https://boardgamearena.com/gamepanel?game=concept}}. For each language, we collect 100 games, which amounts to circa 1,000 concepts played per language. 
This task aims to answer the following research questions: 
\begin{enumerate}
    \item Compared to humans, how well can LLMs infer concepts from natural language clues, selected by a human from a constrained vocabulary?
    To what extent can LLMs interpret information expressed through linguistic formulations that differ from how such concepts are typically represented in their training data?
    \item Is the performance on Concept language-dependent? How well do models transfer abstractions to lower-resource languages?
\end{enumerate}

We hypothesize to observe the following:
\begin{enumerate}
    \item We expect that model performance does not match, but correlates with human performance. Additionally, we expect that reasoning models outperform non-reasoning models, and that everyday and concrete concepts (e.g., objects, animals) are easier for models compared to infrequent and abstract concepts (e.g., quotes).
    \item Performance on low-resource languages is worse than on English.
\end{enumerate}

Our results show that LLMs struggle on a simple language game that humans solve easily (94\% solve rate for non-expert humans). Specifically, across all tested models, we observe difficulties with (i) interpreting other players' strategic intent from the clues they provide, and (ii) correcting initial hypotheses when new information becomes available.\\
The remainder of the paper is structured as follows. In Section \ref{related work}, we discuss prior literature. Section \ref{methodology} introduces the methodology regarding data collection, model prompting, and evaluation. In Section \ref{results}, we discuss our results on the Concept benchmark\footnote{The data is publicly available for research purposes only: \url{https://huggingface.co/datasets/IneG/concept}.}, and Section \ref{conclusion} concludes this study, summarizing the findings and discussing future works.

\section{Related Work} \label{related work}
As mentioned earlier, the core challenge that is addressed in this work relates to logical inference and abductive reasoning. The guesser must interpret the (incomplete) abstract clues to infer the most plausible concept, relying on their own background knowledge, Theory of Mind of the other player, and previous incorrect guesses.
Language games offer an interesting perspective to test these capabilities in AI models. 
For instance, situated language understanding is evaluated through dialogue games \citep{chalamalasetti2023clembench}, Theory of Mind (ToM) -the capacity to reason about others’ concealed mental states- is tested through collaborative games \citep{li-etal-2023-theory}, physical common sense is probed by asking models to organize rooms and place objects in appropriate locations \citep{zhuo-murata-2024-utilizing}, and world knowledge can be tested through the popular game `Twenty Questions`  \citep{de-bruyn-etal-2022-20q}. 
It has also been shown that LLMs as players do not necessarily match human performance, but their performance correlates strongly with difficulty as indicated by humans \citep{xiao2024llms}.\\ 
An interesting subsection of games that are implemented for LLMs are games in which there is a perfect common interest between players.
In this setup, there are usually two players: the `sender', and `receiver'.
The receiver can observe the signals that are sent, but not the state of the world. A message is encoded in signals by the sender, and the receiver must then try to decode them and decide on which actions to take. For each state, there is only one correct action and both the sender and the receiver have the same outcome objective, namely for the receiver to perform the correct action.
Prior research has presented such games as a method to evaluate automated methods, since they test a system’s ability to strategically encode and/or decode information, align meaning between agents, and coordinate actions within a shared framework.\\
For instance, LLMs have made considerable gains in solving crossword puzzles, especially with the implementation of search algorithms \citep{saha2024language}. However, cryptic puzzles (e.g., relying on wordplay) are much more challenging, and there remains a large performance gap compared to humans \citep{saha2024language,sadallah2024llms}. \citet{leng2025crosswordbench} create a text+grid crossword benchmark, and show that reasoning models outperform non-reasoning models. However, they do not mention a human baseline. Similarly, the game `Codenames' has been introduced as benchmark to test language nuances and ToM, on which LLMs again outperform older methods. However, \citet{stephenson2025codenames} note that smaller models do not adhere to the constraints of game rules, and models struggle overall with lateral thinking. On the popular puzzle `Wordle', LLMs again underperform but CoT-prompting increases performance \citep{xiao2024llms}.
Additionally, the `Connections' puzzle in the New York Times, where players have to make four clusters from sixteen words, is still challenging for LLMs. Especially when the model's first answer is incorrect, the accuracy is low \citep{todd2024missedconnectionslateralthinking}. 
Similarly, in Italian, the game `La Ghigliottina' in which a single player has to determine which concept connects a set of five seemingly unrelated clues, shows that LLMs are still underperforming \citep{manna-etal-2024-riddle}. 

However, these games are based on static situations. Lately, interactive games have been used to gauge models' continuous understanding updates, context-aware responses, and the inference of hidden states of other agents \citep{momente2025triangulating}. These dynamic, or interactive games provide an information exchange between the sender and receiver. It is argued that these games offer a better proxy for the real-world necessity of adaptability and robustness in interactions. Such game-based evaluations have demonstrated higher correlations with a range of cognitive abilities, including causal and logical inference, as well as social and emotional competencies \citep{momente2025triangulating}.
For instance, in `Guess The Build', players construct a Minecraft build and others try to guess it in natural language. Results show a considerable performance gap between open and API Visual Language Models (VLMs). The performance correlates with frequency of the concept in the training data, and underrepresented cultural concepts and low-resource languages remain an open problem \citep{zhu2025guessbenchsensemakingmultimodalcreativity}. Similarly, the game `Pictionary' has been adapted as a multi-modal benchmark `Iconary', in which a guesser tries to identify a concept that is drawn by composing a set of icons, and the drawer can update their drawing to help the guesser. Results show that T5-models are better at guessing than drawing, and the main challenge were out-of-vocabulary words \citep{clark-etal-2021-iconary}. Pictionary has also been played by VLMs using the dataset and agentic framework `Sketchtopia', where the win rate is lower compared to humans, and the performance is again linked to frequency in training data \citep{khan2025sketchtopia}.\\

In order to be successful in such games, being able to estimate mental states of other players (such as their knowledge, beliefs, or desires), i.e., ToM, is important. While \citet{kosinski2024evaluating} observes that these skills are improving with newer and larger LLMs,
\citet{akata2025playing} find that LLMs often fail in game settings where cooperation with other players is required, suggesting that ToM of other players is limited. Similarly, \citet{li2023theory} find long contexts and hallucinations are barriers to ToM in LLMs for multi-agent collaboration in text games. Additionally, \citet{agashe2023llm} show that LLMs "perform well in multi-agent coordination settings where decision-making primarily relies on environmental variables but face challenges in scenarios requiring active consideration of partners’ beliefs and intentions".\\
Concept adds to this line of research by presenting the task both as a static challenge and a dynamic one, bridging the gap between the two usually separate methodologies. In the static setting, we present the final set of clues, whereas in the dynamic setting we update the clues iteratively. Additionally, our setup allows us to compare LLM performance directly to a human baseline, which is often not possible for other games. Moreover, whereas the majority of benchmarks and games are focused on English, our benchmark covers several languages, allowing us to assess abductive reasoning in a genuinely multilingual setting. The core challenge in Concept, inferring the most plausible answer given ambiguous/ incomplete set of hints, has been tackled before in a multimodal approach (e.g., `Guess The Build' and `Pictionary'), but to the best of our knowledge not in a \textbf{single-modal} and \textbf{multilingual} setting. This makes the task more straightforward, and aligns the task representation more with the LLMs' pre-training data.

\section{Methodology} \label{methodology}
\subsection{Data}
We collected game logs of Concept from the online boardgame platform Board Game Arena (BGA)\footnote{\url{https://en.boardgamearena.com/gamepanel?game=concept}. We will release the data for research purposes only.}. Only completed games were kept, and all player data was anonymized (i.e., we removed player ID, timing, stats, etc.). For each language in our dataset (English, French, Dutch, and Spanish), we collected 100 game logs, which amounts to circa 1,000 concepts played per language. In this section, we will detail the structure of our dataset.\\
Each game includes multiple rounds, with one concept per round. On average, a single game consists of 12 concepts. While each concept within a game is unique, the same concept can appear across different games; in our dataset, the most frequent concept occurs nine times within one language. Importantly, even when the target concept is repeated, the clues differ, since they are provided by different players in different games.\\
Player 1’s clues are structured hierarchically. The opening clue establishes the top-level category (e.g., man, object, quote), and subsequent clues may either refine this category or form nested sub-hierarchies. In the game interface, these hierarchies are visually marked with colors: each color contains a distinct high-level clue (a pawn), with its own set of subordinate clues (cubes). Player 1 can choose to add extra clues, or remove clues that were given earlier from the board.
For each round, we collected (i) the concept; (ii) the clues that were given by Player 1 (with hierarchical structure indicated by color); (iii) the guesses made by Player 2.

On the collected dataset, we filter out the rounds for which no clues were given (i.e., Player 2 guessed the concept before Player 1 gave any clues). In addition, some rounds contain very few clues because Player 1 removed most of them at the end of the game. To account for this, we report results on two versions of the dataset: one containing all rounds, and one excluding incomplete rounds. Our filtering heuristic considers a round incomplete if the initial clue (the top-level and most important clue) is missing at the end of the game. Table \ref{Table:dataset-stats} provides an overview of the resulting dataset sizes.

\begin{table}[!t]
  \centering
  \resizebox{\columnwidth}{!}{%
    \begin{tabular}{||c| c c c ||} 
      \hline
      & excluded rounds & total rounds & filtered rounds \\ [0.5ex] 
      \hline\hline
      English   & 72 & 1,103 & 818\\ 
      French  & 89 & 1,036 & 829\\
      Dutch & 40 & 1,155 & 932\\
      Spanish & 28 & 924 & 716\\ [1ex] 
      \hline
    \end{tabular}%
  }
  \caption{Dataset sizes: excluded rounds (no clues given); total number of rounds (rounds remaining after excluding empty clues); filtered rounds (only complete rounds, in which the first clue is still present at the end of the game).}
  \label{Table:dataset-stats}
\end{table}

\begin{figure}
    \centering
    \includegraphics[width=1\linewidth]{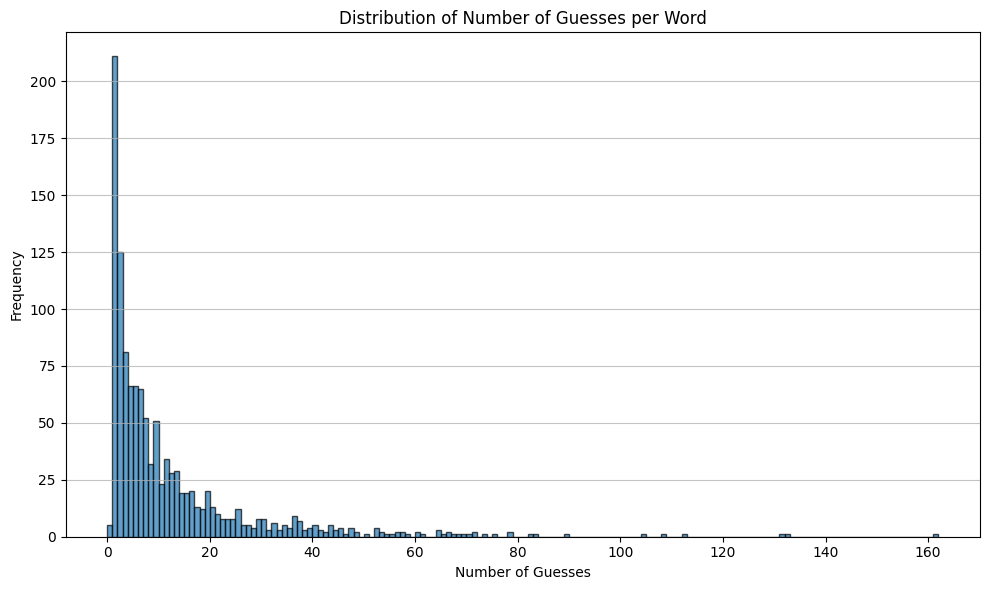}
    \caption{Distribution of number of guesses by humans.}
    \label{fig:nr_guesses_human}
\end{figure}

There is a large variety in how many guesses humans made to either find the correct word or give up, as can be seen in Figure \ref{fig:nr_guesses_human}. The majority of the words only needed <10 guesses.

\subsection{Prompting}
We include a wide range of LLMs, both open- and closed-sourced: LlaMA3.3 70B; LlaMA4-Maverick; Qwen2.5 72B; Deepseek V3; Mistral Small 24B; and GPT 4.1 mini. Additionally, we include the GPT-OSS-120B reasoning model. These models were chosen because they are among the most recent high-performing releases across major model families, spanning diverse architectures and parameter scales, offering a representative overview of current LLM capabilities. In Appendix \ref{appendix-models}, we include more model details and hyperparameter settings.\\
We keep the prompt identical for all models. We organize the clues from Player 1 in a hierarchical structure (grouping clues of the same color together), since tests on the separate validation set (n=50) showed that this technique results in better performance. There is no limit to the number of clues that can be given, as is the case in the original human game.
We also include one example in the prompt, taken from the rule book of the original game. Prior works found that one example can work better than multiple examples, `possibly due to cognitive overload or excessive focus on example analysis instead of task solving' \citep{kaesberg2025sparc,ye2023comprehensivecapabilityanalysisgpt3}.\\

We implement two prompts for the main experiments (in English). Specifically, we create a \textbf{static} and a \textbf{dynamic} prompt. As addressed in Section \ref{related work}, the state of the game can either be given statically, where Player 2 gets all information at once, or dynamically, where the information is updated sequentially and there is an interaction between the two players.\\
The static prompt consists of the final set of clues given by Player 1 (similar to a snapshot of the clues as they were at the end of the round), and any incorrect guesses made by the LLM are added to the prompt. We allow 10 guesses per round. This informs us on how well LLMs can navigate the conceptual space, created by humans, and how well they can combine symbolic clues to infer the correct answer. We formalize the procedure in pseudocode \ref{algorithm:static}:

\begin{algorithm}[ht]
\caption{Static prompt}
\label{algorithm:static}
\KwIn{$C$ -- fixed set of clues}
$W \leftarrow [\ ]$\Comment*[r]{incorrect guesses}

\While{true}{
    $answer \leftarrow \text{GENERATE\_ANSWER}(C + W)$\;
    \If{\text{IS\_CORRECT}$(answer)$}{
        \Return $answer$\;
    }
    $W.\text{append}(answer)$\;
}
\end{algorithm}
However, as indicated by prior research, including the interaction between players gives more insights into Theory of Mind capabilities. Specifically, in Concept, Player 2 should keep in mind the strategic intent of Player 1's choice of clues, and why Player 1 might choose to update their clues given Player 2's incorrect answers. Therefore, in the dynamic prompt, we prompt iteratively. Specifically, we update the clues following the original game play. We present the clues per step (one step = the clues given until the original (human) Player 2 interrupted Player 1 by making a guess). For each step, we format the clues as mentioned earlier, and allow the model to guess as many times as humans did at that step. Then, instead of only adding the LLM's incorrect answers, we also provide the human's incorrect answers to approximate the original interactive process. Then, if the LLM has not found the correct answer, we move to the next step, updating the clues, and repeating the process. We formalize the procedure in pseudocode \ref{algorithm:dynamic}:

\begin{algorithm}[ht]
\caption{Dynamic prompt}
\label{algorithm:dynamic}
\KwIn{$C_1,\dots,C_T$ -- clue sets for steps $t = 1,\dots,T$}
\KwIn{$H_1,\dots,H_T$ -- human incorrect answers at each step (lists)}
$W \leftarrow [\ ]$\Comment*[r]{global list of incorrect answers}
\For{$t \leftarrow 1$ \KwTo $T$}{
    $W.\text{extend}(H_t)$\Comment*[r]{add human wrong answers at step $t$}
    \For{$k \leftarrow 1$ \KwTo $\lvert H_t\rvert$}{
        $answer \leftarrow \text{GENERATE\_ANSWER}(C_t + W)$\;
        \If{\text{IS\_CORRECT}$(answer)$}{
            \Return $answer$\;
        }
        $W.\text{append}(answer)$\Comment*[r]{record model's wrong answer}
    }
}
\end{algorithm}

For the multilingual prompting, we translated the English prompt to the target language\footnote{Each translation was proofread by at least a CEFR B2 (ACTFL Advanced Mid) speaker.}. We compared translating the entire prompt to the target language, and keeping the instructions in English on a validation set \citep{vatsal2025multilingualpromptengineeringlarge}. Given that the latter yielded better results, on the test set we prompt with the entire prompt in the target language. We include all prompts in Appendix \ref{appendix-prompt}.

\subsection{Evaluation}
We evaluate model performance using precision at k (prec@k), where a prediction is considered correct only if the guessed concept exactly matches the target concept. To ensure fair comparison, we apply extensive normalization to the LLM outputs, including correcting misrepresented characters (e.g., ASCII artifacts in non-English concepts), and cleaning responses such as removing the prefix `**Concept**:' when present. 
For the static prompt setting, we fix $k$ at 10. For the dynamic prompt setting, $k$ is set to the number of guesses made by humans in the corresponding step (see Figure \ref{fig:nr_guesses_human}).

\section{Results} \label{results}

\begin{table*}[!ht]
  \centering
  \resizebox{\textwidth}{!}{%
    \begin{tabular}{||c| c c c c c c c c||} 
      \hline
      & Human & llama3-70B & Qwen2.5-72B & Deepseek-V3 & llama4-maverick & Mistral-24B & GPT-4.1 mini & GPT-OSS-120B\\ [0.5ex] 
      \hline\hline
      Aggregated (static)   & 0.9184 & 0.3469 & 0.2982 & 0.3463 & 0.2747 & 0.2955 & 0.3445 & 0.2792\\ 
      Aggregated (dynamic)  & 0.9184 & 0.2533 & 0.2336 & 0.3054 & 0.2246 & 0.2075 & 0.2893 & 0.2318\\
      Only complete (static) & 0.9339 & 0.3924 & 0.3496 & 0.3863 & 0.3117 & 0.3337 & 0.3728 & 0.3129\\
      Only complete (dynamic)& 0.9339 & 0.2713 & 0.2603 & 0.3190 & 0.2432 & 0.2139 & 0.3068 & 0.2493\\ [1ex] 
      \hline
    \end{tabular}%
  }
  \caption{Found rates per model on the English dataset. Entire dataset: n=1,103, filtered: n=818.}
  \label{Table:found-rates}
\end{table*}

\begin{table*}[!h]
  \centering
  \resizebox{\textwidth}{!}{%
    \begin{tabular}{||c|l| c c c c c c c c||}
      \hline
      Language & Split & Human & llama3-70B & Qwen2.5-72B & Deepseek-V3 & llama4-maverick & Mistral-24B & GPT-4.1 mini & GPT-OSS-120B \\ [0.5ex]
      \hline\hline
      \multirow{2}{*}{French}
        & Aggregated (static)    & 0.9403 & 0.1462 & 0.1583 & 0.2123 & 0.2123 & 0.1438 & 0.2133 & 0.2133 \\
        & Only complete (static) & 0.9457 & 0.1712 & 0.1845 & 0.2460 & 0.2448 & 0.1640 & 0.2376 & 0.2279 \\ [1ex]
      \hline
      \multirow{2}{*}{Dutch}
        & Aggregated (static)    & 0.9402 & 0.2008 & 0.1696 & 0.2320 & 0.2112 & 0.1350 & 0.2363 &0.2363 \\
        & Only complete (static) & 0.9484 & 0.2167 & 0.1856 & 0.2371 & 0.2210 & 0.1480 & 0.2435 & 0.2414 \\  [1ex]
      \hline
      \multirow{2}{*}{Spanish}
        & Aggregated (static)    & 0.9469 & 0.2208 & 0.2218 & 0.3131 & 0.2622 & 0.2239 & 0.2823 & 0.2845 \\
        & Only complete (static) & 0.9497 & 0.2555 & 0.2486 & 0.3351 & 0.2877 & 0.2513 & 0.3100 & 0.3114 \\ [1ex]
      \hline
    \end{tabular}%
  }
  \caption{Found rates per model for French, Dutch, and Spanish for the static prompt. Entire dataset French: n=1,036, filtered: n=829. Entire dataset Dutch: n=1,155, filtered: n=932. Entire dataset Spanish: n=942, filtered: n=716.}
  \label{Table:found-rates multilingual}
\end{table*}

\subsection{Human performance}
It is clear that Concept is an easy task for humans. To some extent, for persons who play the game very frequently, there might be a learning effect, but since there are 990 unique concepts per language in the game and our dataset includes 100 unique game plays, we can assume this effect is negligible. 
Considering all rounds, humans achieve a 92\% found rate, and 93\% on the filtered rounds. 63\% of the rounds were solved within 10 guesses, although the correct answer was often still reached when more guesses were required. As expected, there was a negative correlation (Pearson's $r = -0.3$) between the number of clues provided and the likelihood of success, which could partially be explained because more difficult concepts requiring more clues.
To examine whether performance varied across types of concepts, we categorized rounds by their first (and most important) clue (e.g., wildlife/animal, object/thing/package, building/construction/city, music/song/note), focusing on the top 20 most frequent categories. A Bayesian mixed-effects logistic regression model (GLMM)\footnote{\url{https://bambinos.github.io/bambi/notebooks/getting_started.html\#generalized-linear-mixed-models}} was then fit with `found' (i.e., whether the concept was found) as the outcome and category as a predictor\footnote{Posterior estimates were obtained with NUTS sampling (4 chains; 1,000 warmup and 1,000 post-warmup draws), and category effects were considered meaningful when their 95\% highest-density intervals excluded 0.}. The analysis revealed no meaningful differences across categories, indicating that human performance was consistently high regardless of concept type. We refer to Table \ref{Table:found-rates per category} in Appendix \ref{appendix:per category} for a complete overview of the results.\\

\subsection{LLM performance}
\begin{figure}
    \centering
    \includegraphics[width=1\linewidth]{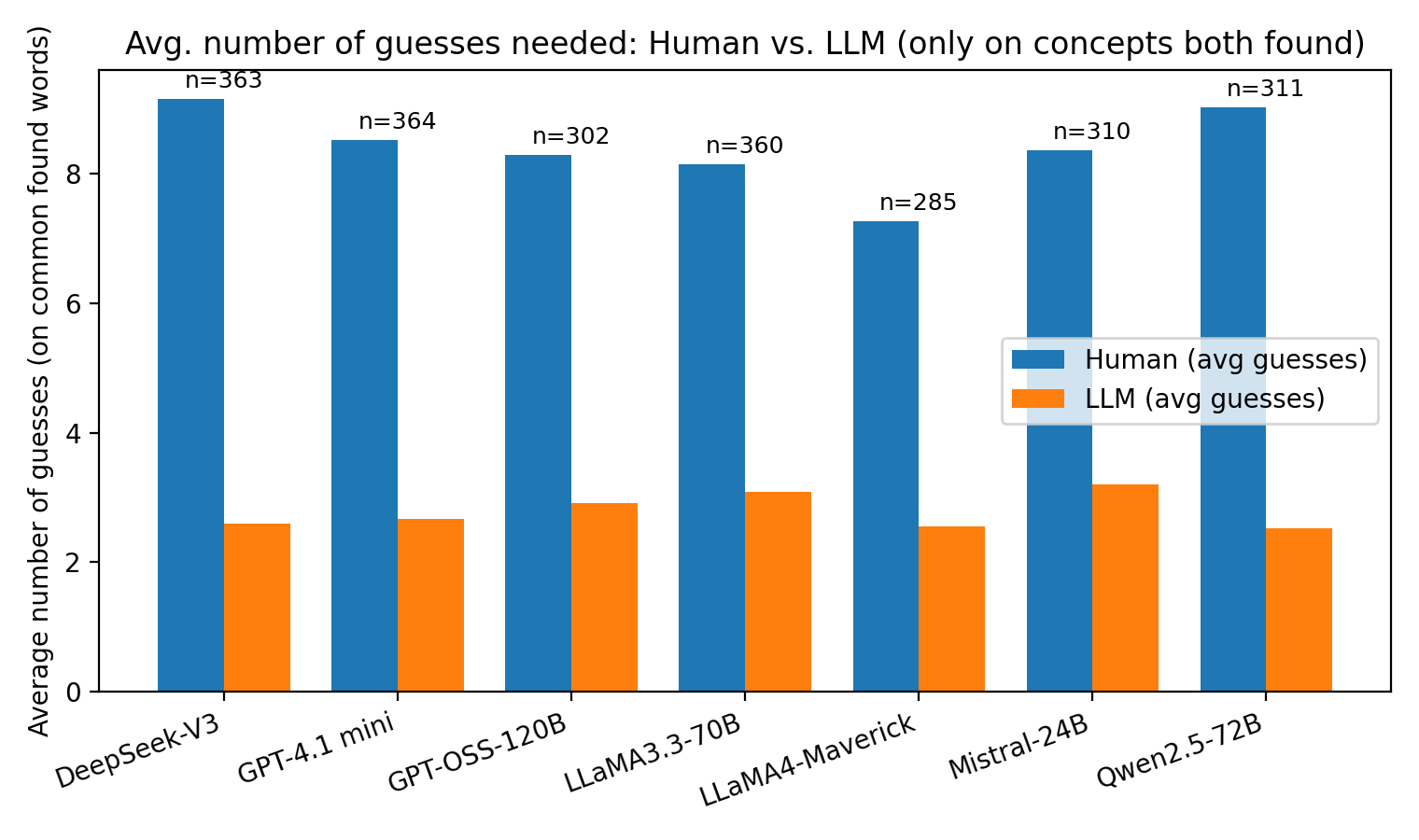}
    \caption{Number of guesses needed to arrive at the correct concept, for concepts that are found both by humans and the LLM.}
    \label{fig:comparison_guesses}
\end{figure}

In comparison, all LLMs perform significantly worse than humans (see Table \ref{Table:found-rates}). Interestingly, in rounds where both humans and the LLM found the correct concept, the LLM overall needs fewer guesses to do so, suggesting greater efficiency once they are on the right track (see Figure \ref{fig:comparison_guesses}). 

However, in rounds where the correct answer is not found easily, LLMs often get stuck pursuing wrong directions, whereas humans are better at revising their earlier interpretations and still arriving at the correct answer\footnote{For models, we observe a strong negative correlation between found rate and number of guesses (Pearson's $r$ between -0.68 and -0.74), whereas for humans the correlation is weak ($r = -0.14$).}. This finding echoes Chollet's \citeyear{chollet2019measureintelligence} view of exploration as a key component of intelligence: humans can flexibly revise their hypotheses, but LLMs tend to remain fixated. Since Concept requires interpreting strategic linguistic signals within a shared conceptual space, success depends not only on linguistic fluency but also on adaptive exploration—an area where humans still hold a clear advantage.

\subsubsection{Correlation with human performance}
To gauge how models compare to each other and to humans on which concepts they (do not) find, we calculate Pearson's $r$ on the found rates, based on the static prompt. We find that GPT-OSS-120, the reasoning model, has the strongest correlation with human found rates ($r=0.14$). Especially LlaMA4-Maverick has a low correlation with human performance ($r= 0.05$). Interestingly, comparing model performance to each other, the correlations are very low, suggesting that different models succeed on largely different subsets of concepts.\\ 
Further, we compare human and LLM performance per category (grouped by the first, highest-level clue), by extending the GLMM model to include solver (human vs. LLM), high-level-clue category, and their interaction as fixed effects. This allowed us to test whether differences in success rates between humans and LLMs varied systematically by concept category. For most models, there is no significant difference, except for Qwen2 72B (worse on `expression/quote, speak, word', and `theater/film/camera'), LlaMA4-Maverick (worse on `expression/quote, speak, word'), and Mistral-Small 24B (worse on `expression/quote, speak, word', and `object/thing/package'). To some extent, this agrees with our initial hypothesis that LLMs would struggle more with infrequent and culturally-dependent concepts, such as expressions and quotes. We refer to Table \ref{Table:found-rates per category} in Appendix \ref{appendix:per category} for a complete overview of the results.

\subsubsection{Static vs. dynamic}
Contrary to our expectations, the dynamic prompt yields worse results compared to the static prompt, even though this prompt entails more information, and provides the added possibility to derive useful information from the interaction between Player 1 and Player 2. Corroborating prior research, this suggests that LLMs struggle with simulating Player 1's strategic intents when giving the clues in in-context learning. Further, as mentioned earlier, even with more explicit representations of the updates in clues and incorrect guesses, LLMs are seemingly unable to update their interpretations of the initial clues, and remain stuck in the wrong direction.\\
A qualitative error analysis shows that both for the static and dynamic setting, models often repeat their previous (incorrect) answer, even though the prompt is updated with this information. Additionally, there are rounds where the guesses are not valid given the constraints of the highest level clue: for instance, if the first clue is `building', the LLM's guesses are not buildings.

\subsubsection{Reasoning vs. non-reasoning models}
Again contrary to our initial expectations, the reasoning model we included in our experiments (GPT-OSS-120B) does not outperform non-reasoning models. In some cases, this is caused because the number of reasoning tokens (n=1,000) were insufficient and the model did not arrive at an answer. This finding agrees with previous results that longer reasoning processes, in which the model generates more intermediate tokens, does not necessarily help the model to arrive at useful associations or correcting the erroneous directions \citep{yuan2025turnaboutllmdeductivereasoningbenchmark,hassid2025dontoverthinkitpreferring,vamvourellis2025reasoningoverthinkingevaluatinglarge}.

\subsection{Multi-lingual performance}
Finally, we observe similar trends for Dutch, French, and Spanish, although the differences are even more pronounced. While human performance remains consistently high (94\% found rate), all LLMs perform worse, and the gap relative to humans is larger than in English (see Table \ref{Table:found-rates multilingual}). Between the non-English languages, models seem to perform better on Spanish. This could be a reflection of common internet language distributions, where Spanish has a larger share compared to the other two languages. Unfortunately, however, the models' documentation does not provide detailed frequency distributions for non-English training data, so this remains a hypothesis. Most model specifications list French and Spanish as supported languages, but not Dutch, yet LLMs' performance on Dutch is not significantly worse than on French.
Overall, the weak performance of LLMs on non-English Concept games highlights their limited capacity to transfer knowledge for this type of abstraction.

\section{Conclusion} \label{conclusion}
In this paper, we present the word guessing game Concept as a test for abductive reasoning. Comparably to a single-modal version of Pictionary, Player 1 selects clues from a predefined set to describe a concept, Player 2 has to guess this concept. Whereas similar game-related benchmarks have addressed shortcomings in LLMs' reasoning capacities, Concept levels the playing field by having the task representations rely only on natural language, which is better aligned with the pre-training data of LLMs.\\
We collected game logs in English, Dutch, French, and Spanish from the online boardgame platform Board Game Arena, providing a valuable human baseline performance.\\
To showcase the applicability of Concept, we provide baselines from 7 state-of-the-art LLMs, both open- and closed-source.
This language game, which is very easy for humans (with success rates over 90\%), reveals important shortcomings of current LLMs. Specifically, we confirm previous findings that LLMs struggle to interpret Player 1's strategic intents when giving the clues, and fail to update their initial interpretations of the clues when sequential information updates are given. Even though incorrect guesses made by the model are added to the prompt, this does not help the model to arrive at the correct answer: often, the model remains stuck in the wrong direction. Additionally, we noted that in cases where both the LLM and humans identified the concept, LLMs consistently require fewer guesses, but in cases where more guesses are needed, humans are better at correcting their abstractions to still find the correct answer.\\
Comparing results from the dominant English language to lower-resource languages (Spanish, French, and Dutch), we observed that LLMs perform worse on non-English languages.\\
For further research, we suggest fine-tuning an LLM for this task. Further, for interpretability, we suggest investigating the link between found rate of a concept and its frequency in the LLM's pre-training data, or singling out the importance of one clue by implementing an ablation study. Additionally, since Concept can be played with multiple people guessing, it would be interesting to implement this as a multi-agent setting, in which multiple models guess, and each model sees the output of the others. Given the low correlation between models' found rates, the interaction between different models could potentially aid performance.

\section*{Limitations}
This study is subject to a few limitations. First of all, our evaluation metric, Exact Match, might miss more fine-grained insights. For instance, synonyms of the correct answer are now marked as incorrect. Nonetheless, this conservative measure provides a clear, reproducible benchmark. Moreover, even under this strict criterion, humans still achieve a success rate of over 90\%, underscoring that the task remains feasible.
Second, we observed instances where the LLM did not adhere to the instructions of only outputting the final answer, and is interrupted mid-stream due to the limited number of allowed output-tokens. Since the model was instructed to return only the answer as output, we rule this behavior a task failure, and count these cases as incorrect. 
Third, prompting various models inevitably introduces a trade-off between reproducibility and model sensitivity. We chose to maintain an identical prompt across all models to ensure stability and reproducibility, but we acknowledge that different models might have performed better with variations of the prompt.

\section*{Acknowledgments}
This research was made possible with a grant from the Fonds Wetenschappelijk Onderzoek (FWO) project 42/FA030100/9770. We would like to thank Victor De Marez for his practical support in retrieving the game logs from the Board Game Arena platform. Also many thanks to Pieter Fivez and Luna De Bruyne for insightful brainstorming and discussions.

\bibliography{custom}

\appendix

\section{Model specifications} \label{appendix-models}
In Table \ref{tab:model_specifications}, we detail model name, version, and release date of the models included in our experiments. For non-reasoning models, we allow 10 output tokens, and keep the temperature stable at 0.1 to keep the output more deterministic and reproducible. For GPT-OSS-120B, we allow 1,000 output tokens, with medium reasoning effort, and follow the recommended temperature setting (t=1). Because of hardware limitations, we prompt the models using the Together.ai API\footnote{\url{https://www.together.ai/}}, except for GPT 4.1 mini, for which we use the OpenAI API\footnote{\url{https://platform.openai.com/docs/models/gpt-4.1-mini}}. The total cost for prompting the models was 120\$.

\begin{table*}[!h]
\centering
\scriptsize % makes table more compact
\renewcommand{\arraystretch}{1.2}
\begin{tabular}{llllP{0.36\textwidth}}
\hline
\textbf{Model} & \textbf{Full Name / Variant} & \textbf{Size} & \textbf{Release Date} & \textbf{Source} \\
\hline
LLaMA-4 Maverick & LLaMA-4-Maverick-17B-128E-Instruct-FP8 & 17B active (400B total) & Apr 2025 & \url{https://api.together.ai/models/meta-llama/Llama-4-Maverick-17B-128E-Instruct-FP8} \\
LLaMA-3.3 & LLaMA-3.3-70B-Instruct-Turbo & 70B & Dec 2024 & \url{https://api.together.ai/models/meta-llama/Llama-3.3-70B-Instruct-Turbo} \\
Qwen-2.5 & Qwen-2.5-72B-Instruct-Turbo & 72B & Sept 2024 & \url{https://api.together.ai/models/Qwen/Qwen2.5-72B-Instruct-Turbo} \\
DeepSeek V3 & DeepSeek-V3 & 671B total, 37B active & Dec 2024 & \url{https://api.together.ai/models/deepseek-ai/DeepSeek-V3} \\
Mistral Small & Mistral-Small-24B-Instruct-2501 & 24B & Jan 2025 & \url{https://api.together.ai/models/mistralai/Mistral-Small-24B-Instruct-2501} \\
GPT-4.1 mini & gpt-4.1-mini-2025-04-14 & n/a & Apr 2025 & \url{https://platform.openai.com/docs/models/gpt-4.1-mini} \\
GPT-OSS 120B & GPT-OSS-120B (Reasoning) & 120B & Aug 2025 & \url{https://api.together.ai/models/openai/gpt-oss-120b} \\
\hline
\end{tabular}
\caption{Models evaluated in our experiments, with full names, sizes, release dates, and sources. Sizes are as reported in model documentation (for MoE models, both total and active parameters are listed).}
\label{tab:model_specifications}
\end{table*}

\section{Prompt} \label{appendix-prompt}
Below, we include the template for the \textbf{static} prompt in \textbf{English}:\\

\begin{lstlisting}
[INST]Let's play Concept! I will give you clues, you have to guess the concept. Only output the concept, nothing else!
The hints are organized by color. Per color, the hierarchy of the clues is indicated by 'High-level' and 'Low-level'. Low-level clues are given in function of the high-level clue of the same color. I will also give you the incorrect guesses, so you can learn from them. Here is an example: **Hints**:
    We are looking for a concept in this category: Building, Construction, City (highest-level hint, color: green).
    An additional hint in color green is: Metal.
    A high-level hint in color red is: Location, Country, Flag.
    An additional hint in color red is: Blue.
    An additional hint in color red is: White.
    An additional hint in color red is: Red.
    **Concept**: Eiffel Tower

Here are the clues: **Hints** [insert clues here]
**Wrong guesses** [insert previous wrong answers here]
**Concept**[/INST]
\end{lstlisting}

Here, we include the template for the \textbf{dynamic} prompt in \textbf{English}. The main difference with the static prompt is that only clues until a certain step are given, instead of the clues as they appear at the end of a round. Specifically, we define one step as the clues given by Player 1 until Player 2 interrupts by making a guess. For the next step, we update the clues as Player 1 did (adding or removing clues) until Player 2 interrupted again, and so forth. Second, we include both the incorrect guesses made by humans, as well as LLMs for that step.\\

\begin{lstlisting}
[INST]Let's play Concept! I will give you clues, you have to guess the concept. Only output the concept, nothing else!
The hints are organized by color. Per color, the hierarchy of the clues is indicated by 'High-level' and 'Low-level'. Low-level clues are given in function of the high-level clue of the same color. I will also give you the incorrect guesses, so you can learn from them. Here is an example: **Hints**:
    We are looking for a concept in this category: Building, Construction, City (highest-level hint, color: green).
    An additional hint in color green is: Metal.
    A high-level hint in color red is: Location, Country, Flag.
    An additional hint in color red is: Blue.
    An additional hint in color red is: White.
    An additional hint in color red is: Red.
    **Concept**: Eiffel Tower

Here are the clues: **Hints** [insert clues until the current step here]
**Wrong guesses** [insert previous wrong answers from LLMs and humans until the previous step here]
**Concept**[/INST]
\end{lstlisting}

The \textbf{static} prompt in \textbf{Spanish}:\\

\begin{lstlisting}
[INST]¡Juguemos a Concept! Te daré pistas en español y tú tendrás que adivinar el concepto en español. ¡Solo da el concepto, nada más!
Las pistas están clasificadas por colores. Para cada color, la jerarquía de las pistas se indica con 'Nivel alto' o 'adicional'. Las pistas adicionales se dan en función de las pistas de alto nivel del mismo color. También te daré las respuestas incorrectas para que puedas aprender de ellas.
Aquí tienes un ejemplo: **Pistas**:
    Buscamos un concepto en esta categoría: Edificio, Construcción, Ciudad (pista de nivel superior, color: verde).
    Una pista adicional en color verde es: Metal.
    Una pista de alto nivel en color rojo es: Ubicación, País, Bandera.
    Una pista adicional en color rojo es: Azul.
    Una pista adicional en color rojo es: Blanco.
    Una pista adicional en color rojo es: Rojo.
    **Concepto**: Torre Eiffel.

Estas son las pistas: **Pistas** [insert clues here]
**Respuestas incorrectas**: [insert previous wrong answers here]
**Concepto**[/INST]
\end{lstlisting}

The \textbf{static} prompt in \textbf{French}:\\
\begin{lstlisting}
[INST]Jouons à Concept! Je vais vous donner des indices en français, vous devez deviner le concept en français. Ne donnez que le concept, rien d'autre !
Les indices sont classés par couleur. Pour chaque couleur, la hiérarchie des indices est indiquée par 'Haut niveau' et 'Supplémentaire'. Les indices supplémentaires sont donnés en fonction de l'indice de haut niveau de la même couleur. Je vous donnerai également les réponses incorrectes, afin que vous puissiez en tirer des leçons. 
Voici un exemple: **Indices**:
    Nous recherchons un concept dans cette catégorie: Bâtiment, Construction, Ville (indice de plus haut niveau, couleur : vert).
    Un indice supplémentaire de couleur vert est: Métal.
    Un indice de haut niveau de couleur rouge est: Lieu, Pays, Drapeau.
    Un indice supplémentaire de couleur rouge est: Bleu.
    Un indice supplémentaire de couleur rouge est: Blanc.
    Un indice supplémentaire de couleur rouge est: Rouge.
    **Concept**: Tour Eiffel 

Voici les indices: **Indices** [insert clues here]
Réponses incorrectes**: [insert previous wrong answers here]
**Concept**[/INST]
\end{lstlisting}

The \textbf{static} prompt in \textbf{Dutch}:\\
\begin{lstlisting}
[INST]Laten we Concept spelen! Ik ga je aanwijzingen geven in het Nederlands, je moet het concept in het Nederlands raden. Geef alleen het concept, niets anders!
De hints zijn gecategoriseerd op kleur. Voor elke kleur is de hiërarchie van de hints aangegeven door 'Hoog niveau' of 'Extra'. De extra hints worden gegeven in functie van de hints op hoog-niveau van dezelfde kleur. Ik zal je ook de foute antwoorden geven, zodat je hiervan kunt leren. 
Hier is een voorbeeld: **Hints**:
    We zoeken een concept in deze categorie: Gebouw, Bouw, Stad (hoogste niveau hint, kleur: groen).
    Een extra hint in kleur groen is: Metaal.
    Een hint op hoog niveau in kleur rood is: Locatie, Land, Vlag.
    Een extra hint in kleur rood is: Blauw.
    Een extra hint in kleur rood is: Wit.
    Een extra hint in kleur rood is: Rood.
    **Concept**: Eiffeltoren

Hier zijn de hints: **Hints** [insert clues here]
**Foute antwoorden**: [insert previous wrong answers here]
**Concept**[/INST]
\end{lstlisting}

\section{Comparison per category} \label{appendix:per category}
In Table \ref{Table:found-rates per category} we give an overview of the results for each model per semantic category. The categories are defined as the highest-level (first) clue from the game, we analayze the 20 most frequent categories.

\begin{table*}[!h]
  \centering
  \resizebox{\textwidth}{!}{%
    \begin{tabular}{||c| c c c c c c c c||} 
      \hline
      & Human & llama3-70B & Qwen2.5-72B & Deepseek-V3 & llama4-maverick & Mistral-24B & GPT-4.1 mini & GPT-OSS-120B\\ [0.5ex] 
      \hline\hline
      Wildlife, animal (n=83)   & 0.9759 & 0.5662 & 0.5903 & 0.5421 & 0.4337 & 0.5542 & 0.5180 & 0.3855\\ 
      
      Theater, film, camera (n=78) & 0.9358 & 0.3974 & 0.1282 & 0.2948 & 0.2051 & 0.2307 & 0.2564 & 0.1794\\
      
      Expression/quote, speak, word (n=69) & 0.8550 & 0.1594 & 0.0724 & 0.2318 & 0.0579 & 0.1014 & 0.1304 & 0.2028\\

      Object, thing, package (n=57) & 0.9824 & 0.4561 & 0.4210 & 0.3684 & 0.3508 & 0.2280 & 0.4736 & 0.3333\\

      Food, nutrition, edible (n=47) & 0.9361 & 0.4893 & 0.4042 & 0.4468 & 0.3829 & 0.4893 & 0.5319 & 0.4893\\

      Male/man, husband, masculine (n=43) & 0.9302 & 0.2558 & 0.2790 & 0.2558 & 0.2093 & 0.3023 & 0.2790 & 0.3023\\

      Fictional, imaginary, wish (n=28) & 0.8214 & 0.3571 & 0.3928 & 0.3928 & 0.3928 & 0.3928 & 0.5357 & 0.3571\\

      Building, construction, city (n=27)& 0.9259 & 0.3333 & 0.4814 & 0.4814 & 0.2592 & 0.3703 & 0.3333 & 0.2222\\

      Literature, writing, book (n=24) & 0.9583 & 0.3333 & 0.2916 & 0.3333 & 0.2916 & 0.2916 & 0.4583 & 0.2500\\

      Television, broadcast, series (n=22) & 0.9545 & 0.3181 & 0.2727 & 0.4545 & 0.3181 & 0.2727 & 0.3181 & 0.4090\\

      Person, family, group (n=20) & 0.9500 & 0.3000 & 0.1500 & 0.2500 & 0.2000 & 0.1000 & 0.2000 & 0.3000\\

      Female/woman, wife, feminine (n=18) & 0.9444 & 0.3888 & 0.2777 & 0.3888 & 0.2777 & 0.3333 & 0.3333 & 0.3888\\

      Games, toys, youth (n=16) & 0.8750 & 0.5000 & 0.4375 & 0.6250 & 0.5625 & 0.3750 & 0.56250 & 0.5000\\

      Flora, plant, nature (n=15) & 1.000 & 0.4000 & 0.3333 & 0.2666 & 0.3333 & 0.4000 & 0.4000 & 0.3333\\

      Location, country, flag (n=13) & 0.9230 & 0.3846 & 0.4615 & 0.4615 & 0.3846 & 0.3076 & 0.3076 & 0.3846\\

      Water, liquid, aquatic (n=11) & 0.9090 & 0.5454 & 0.5454 & 0.5454 & 0.5454 & 0.4545 & 0.4545 & 0.4545\\

      Music, song, note (n=11) & 0.9090 & 0.0909 & 0.0909 & 0.0909 & 0.0909 & 0.0000 & 0.0909 & 0.1818\\

      Work, profession, craft (n=11) & 1.000 & 0.4545 & 0.6363 & 0.4545 & 0.5454 & 0.3636 & 0.1818 & 0.0909\\

      Tools, construction (n=11) & 1.000 & 0.5454 & 0.4545 & 0.7272 & 0.3636 & 0.7272 & 0.8181 & 0.0909\\ [1ex] 
      \hline
    \end{tabular}%
  }
  \caption{Found rates per category (for each category with over 10 observations) for each model. We report this for the static prompt, on the filtered dataset (n=818) since this is the best setting across all models.}
  \label{Table:found-rates per category}
\end{table*}

\end{document}